\RequirePackage{snapshot}
\documentclass[10pt,twocolumn,letterpaper]{article}

\usepackage{cvpr}
\usepackage{times}
\usepackage{epsfig}
\usepackage{graphicx}
\usepackage{amsmath}
\usepackage{amssymb}

\usepackage[utf8]{inputenc} 
\usepackage[T1]{fontenc}    
\usepackage{url}            
\usepackage{booktabs}       
\usepackage{amsfonts}       
\usepackage{nicefrac}       
\usepackage{microtype}      
\usepackage{cite}  
\usepackage{color, colortbl}
\usepackage{atbegshi,picture}	

\usepackage[pagebackref=true,breaklinks=true,letterpaper=true,colorlinks,bookmarks=false]{hyperref}

\cvprfinalcopy

\ifcvprfinal\fi

\newcommand{\tableSize}[0]{\small}
\newcommand{\MODEL}[0]{3D Mask R-CNN} 
\definecolor{Gray}{gray}{0.9}
\newcolumntype{g}{>{\columncolor{Gray}}c}

\begin{document}

\title{Detect-and-Track: Efficient Pose Estimation in Videos}

\author{
Rohit Girdhar$^{1\thanks{Work done as a part of RG's internship at Facebook}}$ \quad
Georgia Gkioxari$^{2}$ \quad
Lorenzo Torresani$^{2,3\footnotemark[1]}$ \quad
Manohar Paluri$^{2}$ \quad
Du Tran$^{2}$
\\
$^{1}$The Robotics Institute, Carnegie Mellon University
\quad
$^{2}$Facebook
\quad
$^{3}$Dartmouth College \\
\small{\url{https://rohitgirdhar.github.io/DetectAndTrack}}
}

\maketitle

\begin{abstract}
This paper addresses the problem of estimating and tracking human body keypoints in complex, multi-person video. 
We propose an extremely lightweight yet highly effective approach that builds upon
the latest advancements in human detection~\cite{he2017mask} and
video understanding~\cite{carreira2017quo}. Our method operates in two-stages: keypoint estimation in frames or short clips, followed by lightweight tracking to generate keypoint predictions linked over the entire video. For frame-level pose estimation we experiment with Mask R-CNN, as well as our own proposed 3D extension of this model, which leverages temporal information over small clips to generate more robust frame predictions.
We conduct extensive ablative experiments on the newly released multi-person video pose estimation benchmark, PoseTrack, to validate various design choices of our model. Our approach achieves an accuracy of 55.2\% on the validation and 51.8\% on the test set using the Multi-Object Tracking Accuracy (MOTA) metric, and 
achieves state of the art performance on the ICCV 2017 PoseTrack keypoint tracking challenge~\cite{posetrack_challenge}.

\end{abstract}

\section{Introduction}

In recent years, visual understanding, such as object and scene recognition~\cite{zhou2014learning,he2017mask,ren2015faster,papandreou2017towards}, has witnessed a significant bloom thanks to deep visual representations~\cite{krizhevsky2012imagenet,He15resnet,szegedy2017inception,Simonyan14c}. Modeling and understanding human behaviour in images has been in the epicenter of a variety of visual tasks due to its importance for numerous downstream practical applications. In particular, person detection and pose estimation from a single image have emerged as prominent and challenging visual recognition problems~\cite{lin2014microsoft}. 
While single-image understanding has advanced steadily through the introduction of tasks of increasing complexity, video understanding has made slower progress compared to the image domain. Here, the preeminent task involves labeling whole videos with a single activity type~\cite{ucf101,hmdb51,Karpathy_14,Simonyan_14b,LRCN,tran2015c3d,kay2017kinetics,carreira2017quo,Girdhar_17a_ActionVLAD,TSN2016ECCV,Feichtenhofer16convolutional}. While still relevant and challenging, this task shifts the focus away from one of the more interesting aspects of video understanding, namely modeling the changes in appearance and semantics of scenes, objects and humans over time~\cite{gkioxari15rstar,chao15hico,mallya16actions,Girdhar_17b_AttentionalPoolingAction}.

In this work, we focus on the problem of human pose tracking in complex videos, which entails tracking and estimating the pose of each human instance over time. The challenges here are plenty, including pose changes, occlusions and the presence of multiple overlapping instances. The ideal tracker needs to accurately predict the pose of all human instances at each time step by reasoning about the appearance and pose transitions over time. Hence, the effort to materialize a pose tracker should closely follow the state of the art in pose prediction but also enhance it with the tools necessary to successfully integrate time information at an instance-specific level. 

\begin{figure}
    \centering
    \includegraphics[width=\linewidth]{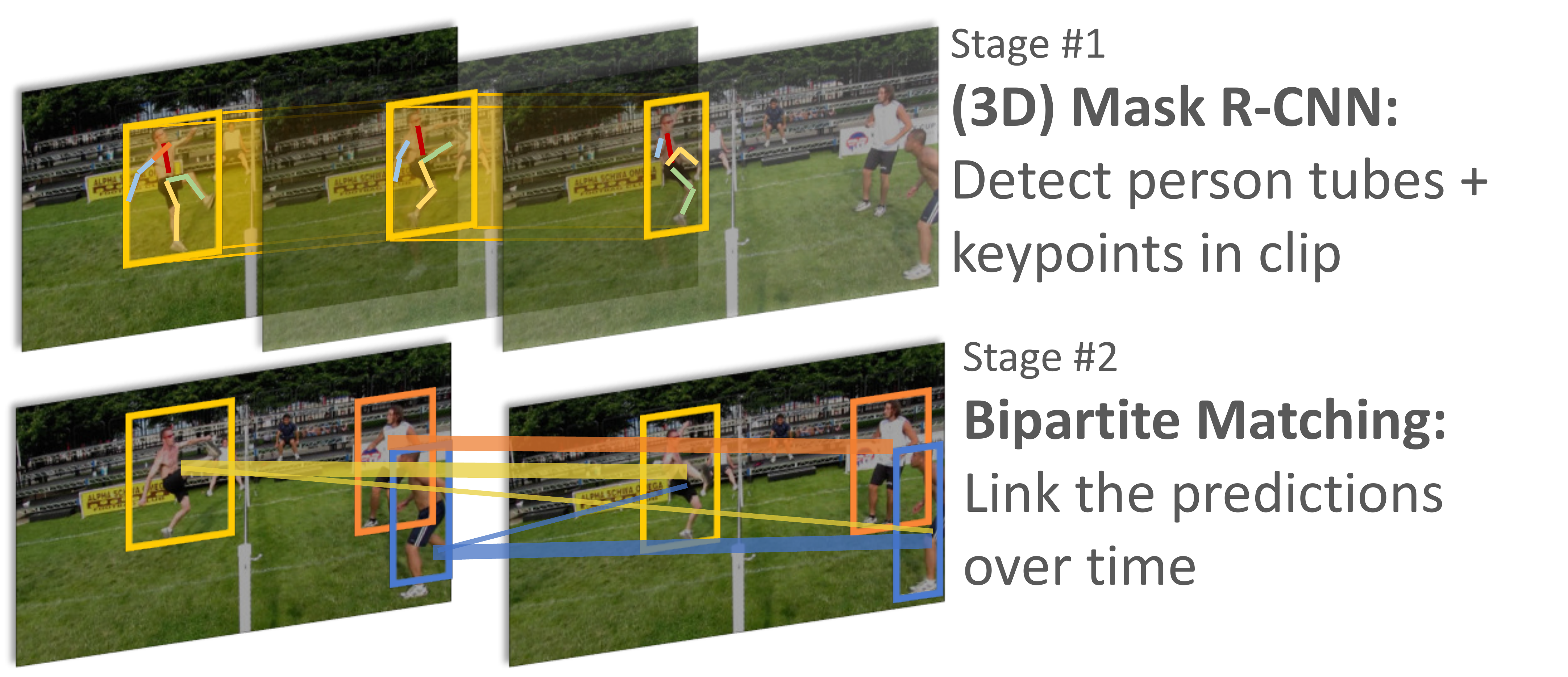}
    \caption{We propose a two-stage approach to keypoint estimation and tracking in videos. For the first stage, we propose a novel video pose estimation formulation, \MODEL{}, that takes a short video clip as input and produces a tubelet per person and keypoints within those. In the second stage, we perform light-weight optimization to link the detections over time.}
    \label{fig:teaser}
\end{figure}

Most recent video pose estimation methods use hand-designed graphical models or integer program optimizations on top of frame-based keypoint predictions to compute the final predictions over time~\cite{insafutdinov2016articulated,iqbal2016pose,song2017thin}. While such approaches have shown good performance, they require hand-coding of optimization constraints and may not be scalable beyond short video clips due to their computational complexity. Most importantly, the tracking optimization is only responsible for linking frame-level predictions, and the system has no mechanism to improve the estimation of keypoints by leveraging temporal information (except~\cite{song2017thin}, though it is limited to the case of single person video). This implies that if a keypoint is poorly localized in a given frame, e.g., due to partial occlusion or motion blur, the prediction cannot be improved despite correlated, possibly less ambiguous, information being at hand in adjacent frames. To address this limitation, we propose a simple and effective approach which leverages the current state of the art method in pose prediction~\cite{he2017mask} and extends it by integrating temporal information from adjacent video frames by means of a novel 3D CNN architecture. It is worth noting that this architecture maintains the simplicity of our two-stage procedure: keypoint estimation is still performed at a frame-level by deploying space-time operations on short clips in a sliding-window manner. This allows our 3D model to propagate useful information from the preceding and the subsequent frames in order to make the prediction in each frame more robust, while using a lightweight module for long-term tracking, making our method applicable to arbitrarily long videos. Fig.~\ref{fig:teaser} illustrates our approach.

We train and evaluate our method on the challenging PoseTrack dataset~\cite{PoseTrack}, which contains real-world videos of people in various everyday scenes, and is annotated with locations of human joints along with their identity index across frames. First, and in order to convince of the efficacy of our method, we build a competitive baseline approach which links frame-level predictions, obtained from Mask R-CNN~\cite{he2017mask}, in time with a simple heuristic. Our baseline approach achieves state of the art performance in the ICCV'17 PoseTrack Challenge~\cite{posetrack_challenge}, proving that it performs competitively on this new dataset. We then propose a 3D extension of Mask R-CNN, which leverages temporal information in short clips to produce more robust predictions in individual frames. For the same base architecture and image resolution, our proposed 3D model outperforms our very strong 2D baseline by 2\% on keypoint mAP and 1\% on the MOTA metric  (details about the metrics in Sec.~\ref{sec:exp:dataset}). In addition, our top-performing model runs at
2 minutes on a 100-frame video, with the tracking itself running in the order of seconds, showing strong potential for practical usage.
As we show in Sec.~\ref{sec:expts:runtime}, this is nearly two orders of magnitude faster than IP based formulations~\cite{iqbal2016pose} using state-of-the-art solvers~\cite{gurobi}.

\section{Related Work}\label{sec:related}

{\noindent \bf Multi-person pose estimation in images:}
The application of deep convolutional neural networks (CNNs) to keypoint prediction~\cite{cao2017realtime,he2017mask,insafutdinov2016deepercut,papandreou2017towards} has led to significant improvements over the last few years. Some of the most recent efforts in multi-person keypoint estimation from still images can be classified into bottom-up versus top-down techniques. Top-down approaches~\cite{he2017mask,papandreou2017towards} involve first locating instances by placing a tight box around them, followed by estimation of the body joint landmarks within that box. On the other hand, bottom-up methods~\cite{cao2017realtime,insafutdinov2016deepercut} involve detecting individual keypoints, and in some cases the affinities between those keypoints, and then grouping those predictions into instances.
Our proposed approach builds upon these ideas by extending the top-down models to the video domain. We first predict spatio-temporal tubes over human instances in the video, followed by joint keypoint prediction within those tubes.

{\noindent \bf Multi-person pose estimation in video:}
Among the most dominant approaches to pose estimation from videos is a two-stage approach, which first deploys a frame-level keypoint estimator, and then 
connects these keypoints in space and time using optimization techniques.
In~\cite{insafutdinov2016articulated,iqbal2016pose}, it is shown that a state of the art pose model followed by an integer programming optimization problem can result in very competitive performance in complex videos. While these approaches can handle both space-time smoothing and identity assignment, they are not applicable to long videos due to the NP-hardness of the IP optimization. Song et al.~\cite{song2017thin} propose a CRF with space-time edges and jointly optimize for the pose predictions. Although they show an improvement over frame-level predictions, their method does not consider body identities and does not address the challenging task of pose tracking. In addition, their approach is hard to generalize to an unknown number of person instances, a number that might vary even between consecutive frames due to occlusions and disocclusions. Our approach also follows a two-stage pipeline, albeit with a much less computationally expensive tracking stage, and is able to handle any number of instances per frame in a video.

{\noindent \bf Multi-object tracking in video:}
There has been significant effort towards multi-object tracking from video~\cite{Reid79,Fortman80}. Prior to deep learning, the proposed solutions to tracking consisted of systems implementing a pipeline of several steps, using computationally expensive hand-crafted features and separate optimization objectives~\cite{yu2016solution} for each of the proposed steps. With the advent of deep learning, end-to-end approaches for tracking have emerged. Examples include~\cite{sadeghian2017tracking,milan2017online} which use recurrent neural networks (RNNs) on potentially diverse visual cues, such as appearance and motion, in order to track objects. In~\cite{Feichtenhofer17DetectTrack}, a tracker is built upon the state of the art object detection system by adding correlation features between pair of consecutive frames in order to predict frame-level candidate boxes as well as their time deformations.
More recent works have attempted to tackle detection and tracking in end-to-end fashion~\cite{hou2017tcnn,kang2016tcnn,kang2016object}, and some works have further used such architectures for down-stream tasks such as action recognition~\cite{hou2017tcnn}.
Our work is inspired by these recent efforts but extends the task of object box tracking to address for the finer task of tracking poses in time.

\section{Technical Approach}\label{sec:approach}

\begin{figure*}
    \centering
    \includegraphics[width=\linewidth]{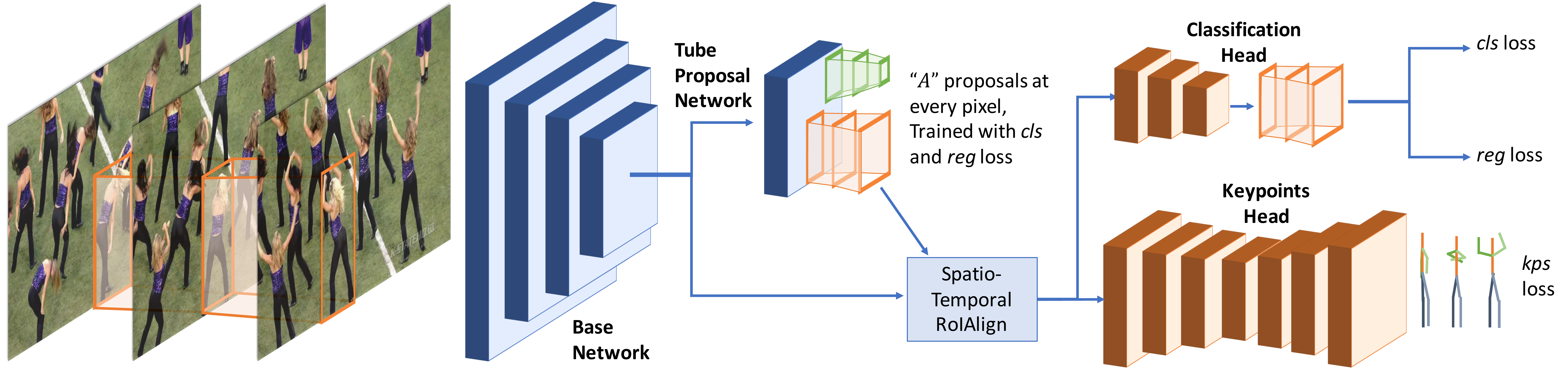}
    \caption{{\bf Proposed \MODEL{} network architecture:} Our architecture, as described in Sec.~\ref{sec:approach:3dmrcnn}, has 
    three main parts. The base network is a standard ResNet, extended to 3D. It generates a 3D feature blob, which is then used to 
    generate proposal tubes using the Tube Proposal Network (TPN). The tubes are used to extract region features from the 3D feature blob,
    using a spatiotemporal RoIAlign operation, and are fed into heads that classify/regress for a tight tube and another to predict keypoint heatmaps.}
    \label{fig:nwarch}
\end{figure*}

In this section, we describe our method in detail. We propose a two-stage approach that efficiently and accurately tracks human instances and their poses across time. We build a 3D human pose predictor by extending Mask R-CNN~\cite{he2017mask} with spatiotemporal operations by inflating the 2D convolutions into 3D~\cite{carreira2017quo}. Our model takes as input short clips and predicts the poses of all people in the clips by integrating temporal information. We show that our 3D model outperforms its 2D frame-level baseline for the task of pose estimation. To track the instances in time, we perform a lightweight optimization that links the predictions. To address exponential complexities with respect to the number of frames in the video and the number of detections per frame, we employ a simple yet effective heuristic. This yields a model that achieves state of the art accuracy on the challenging PoseTrack benchmark~\cite{PoseTrack} and runs orders of magnitude faster than most recent approaches~\cite{insafutdinov2016articulated,iqbal2016pose}.

\subsection{Two-Stage Approach for Pose Tracking}

\paragraph{Stage 1: Spatiotemporal pose estimation over clips.}\label{sec:approach:3dmrcnn}

The first stage in our two-stage approach for human keypoint tracking is pose estimation using a CNN-based model. 
Although our approach can build upon any frame-based pose estimation system, for this work we use Mask R-CNN~\cite{he2017mask} due to its simple formulation and robust performance. Mask R-CNN is a top-down keypoint estimation model that extends the Faster R-CNN object detection framework~\cite{ren2015faster}. It consists of a standard base CNN, typically a ResNet~\cite{He15resnet}, used to extract image features, which are then fed into task-specific small neural networks trained to propose object candidates (RPN~\cite{ren2015faster}), classify them or predict their mask/pose through an accurate feature alignment operation called RoIAlign.

We take inspiration from the recent advancements in action recognition achieved by I3D~\cite{carreira2017quo}, which introduces a video model obtained by converting a state of the art image recognition model~\cite{ioffe2015batch} by inflating its 2D convolutional kernels to 3D.
Analogously, starting from the vanilla Mask R-CNN model, we transform the 2D convolutions to 3D.
Note that the receptive field of these 3D kernels spans over the space and time dimensions and integrates spatiotemporal cues in an end-to-end learnable fashion. Now, the input to our model is no longer a single frame, but a clip of length $T$ composed of adjacent frames sourced from a video. We extend the region proposal network (RPN)~\cite{ren2015faster}, to predict object candidates which track each hypothesis across the frames of the input clip. These tube proposals are used to extract instance-specific features via a spatio-temporal RoIAlign operation. The features are then fed into the 3D CNN head responsible for pose estimation. This pose-estimation head outputs heatmap activations for all keypoints across all frames conditioned on the tube hypothesis. Thus, the output of our \MODEL{} is a set of tube hypotheses with 
keypoint estimates.
Fig.~\ref{fig:nwarch} illustrates our proposed \MODEL{} model, which we describe in detail next.

{\noindent \bf Base network:} We extend a standard ResNet~\cite{He15resnet} architecture to a 3D ResNet architecture
by replacing all 2D convolutions with 3D convolutions. We set the temporal extent of our kernels ($K_T$) to match the
spatial width, except for the first convolutional layer, which uses filters of size $3\times 7\times 7$. We temporally pad the
convolutions as for the spatial dimensions: padding of 1 for $K_T=3$ and 0 for $K_T=1$. We set temporal strides
to 1, as we empirically found that larger stride values lead to lower performance. Inspired by~\cite{carreira2017quo,Feichtenhofer16spatiotemporal},
we initialize the 3D ResNet using a pretrained 2D ResNet. Apart from their proposed
``mean'' initialization, which replicates the 2D filter temporally and divides the coefficients by the number of repetitions, we 
also experiment with a
a ``center'' initialization method, which has earlier been used for action recognition tasks~\cite{Feichtenhofer17spatiotemporal}. In this setup, we initialize the central 2D slice of the 3D kernel with the 2D filter weights and set all the other 2D slices (corresponding to temporal displacements) to zero.
We empirically show in Sec.~\ref{sec:exp:3dmrcnn} that center initialization scheme leads to better performance. The final feature map output of the base 3D network for a $T\times H\times W$
input is $T\times \frac{H}{8}\times \frac{W}{8}$, as we clip the network after the fourth residual block and perform no temporal striding.

{\noindent \bf Tube proposal network:}
We design a tube proposal network inspired by the region proposal network (RPN) in Faster R-CNN~\cite{ren2015faster}.
Given the feature map from the base network, we slide a small 3D-conv network connected to two sibling fully
connected layers -- tube classification ({\em cls}) and regression ({\em reg}). The {\em cls} and {\em reg} labels are
defined with respect to tube anchors. We design the tube anchors to be similar to the bounding box anchors used in Faster R-CNN,
but here we replicate them in time. We use $A$ (typically 12) different anchors at every sliding position, differing in scale and/or aspect ratio. Thus, we have $\frac{H}{8}\times \frac{W}{8}\times A$ anchors in total. For each of these anchors, {\em cls} predicts a binary value indicating whether a foreground tube originating at that spatial position has a high overlap with our proposal tube. 
Similarly, {\em reg} outputs for each anchor a $4T$-dimensional vector encoding  displacements with respect to the anchor coordinates for each box in the tube.
We use the softmax classification loss for training the {\em cls} layer, and the smoothed $L_1$ loss for the  {\em reg} layer. We scale the {\em reg} loss by $\frac{1}{T}$, in order to keep its values comparable to those of the loss for the 2D case. We define these losses as our {\em tracking} loss.

{\noindent \bf \MODEL{} heads:}
Given the tube candidates produced by the tube proposal network, the next step classifies and regresses them into a tight tube around a person track.
We compute the region features for this tube by designing a 3D region transform operator. In particular, we extend the RoIAlign
operation~\cite{he2017mask} to extract a spatiotemporal feature map from the output of the base network.
Since the temporal extent of the feature map and the tube candidates is the same (of dimension $T$), we split the tube
into $T$ 2D boxes, and use RoIAlign to extract a region from each of the $T$ temporal slices in the feature map. The regions are then
concatenated in time to produce a $T\times R\times R$ feature map, where $R$ is the output resolution of RoIAlign
operation, which is kept $7$ for the {\em cls}/{\em reg} head, and $14$ for the keypoint head.
The classification head consists of a 3D ResNet block, similar to the design of the 3D ResNet blocks from the base network; and the 
keypoint head consists of 8 3D conv layers, followed by 2 deconvolution layers to generate the keypoint heatmap
output for each time frame input. The classification head is trained with a softmax loss for the {\em cls} output and a smoothed $L_1$ loss for the {\em reg} output, while the 
keypoint head is trained with a spatial softmax loss, similar to~\cite{he2017mask}.

\paragraph{Stage 2: Linking keypoint predictions into tracks.}\label{sec:approach:stage2}

Given these keypoint predictions grouped in space by person identity (i.e., pose estimation), we need to link them in time 
to obtain keypoint tracks.
Tracking can be seen as a data association problem over these detections. Previous approaches, such as~\cite{pirsiavash2011globally}, have formulated this task as a bipartite matching problem, which can be solved using the Hungarian algorithm~\cite{kuhn1955hungarian} or greedy approaches. More recent work has incorporated
deep recurrent neural networks (RNN), such as an LSTM~\cite{hochreiter1997lstm},
to model the temporal evolutions of features along the
tracks~\cite{milan2017online,sadeghian2017tracking}.
We use a similar strategy, and
represent these detections in a graph, where each detected bounding box (representing a person) in a frame becomes a node.
We define edges to connect each box in a frame to every box in the next frame.
The cost of each edge is defined as the negative likelihood of the two boxes linked on that edge to belong to the same person.
We experimented with both hand-crafted and learned likelihood metrics, which we describe in the next paragraph.
Given these likelihood values, we compute tracks by simplifying the problem to bipartite matching between each pair of adjacent frames.
We initialize tracks on the first frame and propagate the labels forward using the matches,
one frame at a time. Any boxes that do not get matched to an existing track
instantiate a new track. As we show in Sec.~\ref{sec:exp:match-algo}, this simple
approach is very effective in getting good tracks, is highly scalable, is able to deal with a varying number of person hypotheses, and can run on videos of arbitrary length.

{\noindent \bf Likelihood metrics:}
We experiment with a variety of hand-crafted and learned likelihood metrics for linking the tracks.
In terms of hand-crafted features, we specifically experiment with: 1) Visual similarity, defined as
the cosine distance between CNN features
extracted from the image patch represented by the detection; 2) Location similarity, defined
as the box intersection over union (IoU) of the two detection boxes; and 3) Pose similarity,
defined as the PCKh~\cite{yang2013articulated} distance between the poses in the two frames.
We also experiment with a learned distance metric based on a LSTM model that incorporates track history in predicting
whether a new detection is part of the track or not. At test time, the predicted confidence values are used in the matching algorithm, and the matched detection is used to update the LSTM hidden state.
Similar ideas have also shown good performance for traditional tracking tasks~\cite{sadeghian2017tracking}.

In Sec.~\ref{sec:exp} we present an extensive ablative analysis of the various design choices in our two-stage
architecture described above.
While being extremely lightweight and simple to implement, our final model obtains state of
the art performance on the benchmark, out-performing all submissions in the
ICCV'17 PoseTrack challenge~\cite{posetrack_challenge}.

\begin{table*}[t]
  \footnotesize
  \setlength\tabcolsep{2pt} 
  \begin{center}
    \resizebox{\linewidth}{!}{
    \begin{tabular}{lcccccccg|cccccccgccc}
      \toprule
      Threshold &  mAP  & mAP  & mAP  & mAP & mAP & mAP  & mAP   & {\bf mAP}   & MOTA & MOTA & MOTA & MOTA & MOTA & MOTA & MOTA & {\bf MOTA} & MOTP & Prec & Rec  \\
       & Head & Shou & Elbo & Wri & Hip & Knee & Ankl & {\bf Total} & Head & Shou & Elb  & Wri  & Hip  & Knee & Ankl & {\bf Total} & Total& Total& Total\\
      \midrule
      0.0, random tracks & 72.8 & 75.6  & 65.3  & 54.3  & 63.5  & 60.9  & 51.8  & 64.1     & -11.6  & -6.6  & -8.5  & -12.9 & -11.1 & -10.2 & -9.7  & -10.2 & 55.8  & 83.3  & 70.8 \\
      0.0 & 72.8  & 75.6  & 65.3  & 54.3  & 63.5  & 60.9  & 51.8  & 64.1    & 60.3  & 65.3  & 55.8  & 43.5  & 52.5  & 50.7  & 43.9  & 53.6  & 55.7  & 83.3  & 70.8 \\
      0.5 & 72.8  & 75.6  & 65.3  & 54.3  & 63.5  & 60.9  & 51.8  & 64.1    & 61.0  & 66  & 56.3  & 44.1  & 52.9  & 51.1  & 44.3  & 54.2  & 55.7  & 83.3  & 70.8 \\
      0.95 &  67.5  & 70.2  & 62  & 51.7  & 60.7  & 58.7  & 49.8  & 60.6    & 61.7  & 65.5  & 57.3  & 45.7  & 54.3  & 53.1  & 45.7  & {\bf 55.2}  & 61.5  & 88.1  & 66.5 \\
      \bottomrule
    \end{tabular}
    }
  \end{center}
  \caption{{\bf Effect of the detection cut-off threshold}. 
  We threshold the detections computed by Mask R-CNN before matching them to compute tracks. While keypoint mAP goes down, the tracking MOTA performance  increases as there are fewer spurious detections to confuse the tracker. The first row also shows the random baseline; i.e. the performance of the model that randomly assigns a track ID between 0 and 1000 (maximum allowed) to each detection.
  }\label{tab:thresh}
\end{table*}

\section{Experiments and Results}
\label{sec:exp}

We introduce the PoseTrack challenge benchmark and
experimentally evaluate the various design choices of our model.
We first build a strong baseline with our two-stage keypoint tracking approach that obtains
state of the art performance on this challenging dataset. Then, we show how our \MODEL{}
formulation can further improve upon that model by incorporating temporal context.

\subsection{Dataset and Evaluation}\label{sec:exp:dataset}

PoseTrack~\cite{posetrack_data,PoseTrack} is a recently released 
large-scale challenge dataset for human body
keypoint estimation and tracking in diverse, in-the-wild videos. 
It consists of a total of 514 video sequences with 66,374 frames, split
into 300, 50 and 208 videos for training, validation and testing, respectively.
The training videos come with the middle 30 frames densely 
labeled with human body keypoints. The validation and test
videos are labeled at every fourth frame, apart from the middle 30
frames. This helps evaluate the long term tracking performance 
of methods without requiring expensive annotations throughout the entire
video. In total, the dataset contains 23,000 
labeled frames and 153,615 poses.
The test set annotations are held-out, and evaluation are performed
by submitting the predictions to an evaluation server.

The annotations consist of  human head bounding boxes and
15 body joint keypoint locations per labeled person. 
Since all our proposed approaches are top-down and depend on the
detection of the extent of the person before detecting keypoints,
we compute a bounding box by taking the min 
and max extents of labeled keypoints, and dilating that box by 20\%.
Also, to make the dataset compatible with COCO~\cite{coco_dataset,lin2014microsoft},
we permute the keypoint labels 
to match the closest equivalent labels in COCO.
This allows us to pretrain our models on COCO, augmenting the PoseTrack dataset significantly and giving a large improvement in performance.

The dataset is designed to evaluate methods on three different tasks:
1) Single-frame pose estimation; 2) Pose estimation in video; 
3) Pose tracking in the wild. 
Task 1) and 2) are evaluated at a frame level,
using the mean average precision (mAP) metric~\cite{pishchulin2016deepcut}.
Task 3) is evaluated using a multi-object tracking
metric (MOT)~\cite{bernardin2008evaluating}.
Both evaluations require first computing the distance of each
prediction from each ground truth labeled pose. This is done using the
PCKh metric~\cite{andriluka20142d}, which computes the probability
of correct keypoints normalized by the head size.
The mAP is computed as in~\cite{pishchulin2016deepcut},
and the MOT is as described in~\cite{milan2016mot16}.
Their MOT evaluation penalizes false positives equally regardless of their confidence. For this, we drop keypoint predictions with low confidence (1.95 after grid-search on the validation set).
We use the PoseTrack  evaluation toolkit for computing all results presented in this paper,
and report final test numbers as obtained from the evaluation server.

\subsection{Baseline}

In an effort to build a very competitive baseline, we first evaluate the various design elements of our two stage tracking pipeline with a vanilla Mask R-CNN 
base model. This model disregards time-sensitive cues when making pose predictions. 
Throughout this section, our models are initialized from ImageNet and are pretrained on the COCO keypoint detection task.
We then finetune the Mask R-CNN model on PoseTrack, keeping most hyper-parameters fixed to the defaults used
for COCO~\cite{lin2014microsoft}. 
At test time, we run the model on each frame and store the bounding box and keypoint predictions, which are linked
over time in the tracking stage.
This model is competitive as it achieves state of the art results on the PoseTrack dataset. In Sec.~\ref{sec:exp:3dmrcnn}, we prove that our approach can further improve the performance by incorporating temporal context from each video clip via a \MODEL{} model.

{\noindent \bf Thresholding initial detections:}
Before linking the detections in time, we drop the low-confidence and potentially incorrect detections.
This helps prevent the tracks from drifting and reduces false positives. Table~\ref{tab:thresh} shows the effect of thresholding the detections. As expected, the MOTA tracking metric~\cite{bernardin2008evaluating} improves with higher thresholds, indicating better and cleaner tracks.
The keypoint mAP performance, however, decreases by missing out on certain low-confidence detections, which tend to improve the mAP metric. Since we primarily focus on the tracking task, we threshold our detections at 0.95 for our final experiments.

{\noindent \bf Deeper base networks:}
As in most vision problems, we observed an improvement in frame-level pose estimation by using a deeper base model. The improvements also directly transferred to the tracking performance. Replacing ResNet-50 in Mask R-CNN with Resnet-101 gave us about 2\% improvement in MOTA. 
We also observed a gain in performance on using feature pyramid networks (FPN)~\cite{lin2017FPN} in the base network.
Ultimately, we got best performance by using a ResNet-101 model with FPN as the body,
a 2-layer MLP for the classification head, and a stack of eight conv and deconv layers as the keypoint head.

{\noindent \bf Matching algorithm:}\label{sec:exp:match-algo}
We experimented with two bipartite matching algorithms: the Hungarian algorithm~\cite{kuhn1955hungarian} and a greedy algorithm. While the Hungarian algorithm computes an optimal matching given an edge cost matrix, the greedy algorithm takes inspiration from the evaluation algorithms for object
detection and tracking. We start from the highest confidence match, select that edge and remove the two connected nodes out of consideration. This process of connecting each predicted box in the current frame with previous frame is repeatedly applied from the first to the last frame of the video. Table~\ref{tab:match_algo} compares the two algorithms, using the ``bounding box overlap'' as cost metric (details in Sec.~\ref{sec:exp:cost}). We observe that the Hungarian method performs slightly better, thus we use it as our final model.

\begin{table}[t]
  \tableSize{}
  \setlength\tabcolsep{1pt} 
  \begin{center}
    \resizebox{\linewidth}{!}{
    \begin{tabular}{lcccccccgccc}
      \toprule
      Method & MOTA & MOTA & MOTA & MOTA & MOTA & MOTA & MOTA & {\bf MOTA} & MOTP & Prec & Rec  \\
      & Head & Shou & Elb  & Wri  & Hip  & Knee & Ankl & {\bf Total} & Total& Total& Total\\
      \midrule
      Hungarian & 61.7  & 65.5  & 57.3  & 45.7  & 54.3  & 53.1  & 45.7  & {\bf 55.2}  & 61.5  & 88.1  & 66.5 \\
      Greedy & 61.7 & 65.4 & 57.3 & 45.6 & 54.2 & 53 & 45.6 & 55.1 & 61.5 & 88.1 & 66.5 \\
      \bottomrule
    \end{tabular}
    }
  \end{center}
  \caption{{\bf Comparison between Hungarian and Greedy algorithm for matching}. Effect of matching algorithm in tracking performance, computed over the bounding-box overlap cost criterion. The hungarian algorithm obtains slightly higher performance than the simple greedy matching.
  }\label{tab:match_algo}
\end{table}

{\noindent \bf Tracking cost criterion:}\label{sec:exp:cost}
We experimented with three hand-defined cost criteria as well as the learned LSTM metric to
compute the likelihoods for matching.
First, we use bounding box overlap over union (IoU) as the similarity metric.
This metric expects the person to move and deform little from one frame to next, which implies that matching boxes in adjacent frames should mostly overlap. Second, we used pose PCKh~\cite{yang2013articulated} as
the similarity metric, as the pose of the same person is expected to change little between
consecutive frames. 
Third, we used the cosine similarity between CNN features as a similarity metric. In particular, we use the \texttt{res3} layer of a ResNet-18 pretrained on ImageNet, extracted from the
image cropped using the person bounding box. 
Finally, as a learned alternative, we use a LSTM model described in Sec.~\ref{sec:approach:stage2} to learn to match detections to the tracks. Table~\ref{tab:cost} shows that the performance is relatively stable across different
cost criteria. We also experimented with different layers of the CNN, as well as combinations of these cost criteria, all of which performed similarly. 

Given the strong performance of bounding box overlap,
we use the box $x_{min},y_{min},x_{max},y_{max}$ as the input feature for a detection.
Despite extensive experimentation with different LSTM architectures (details in supplementary), we found that the learned metric did not perform as well as the simpler hand-crafted functions,
presumably due the small size of the training set. 
Hence for simplicity and given robust performance, we
use box overlap for our final model.

\begin{table}[t]
  \tableSize{}
  \setlength\tabcolsep{1pt} 
  \begin{center}
    \resizebox{\linewidth}{!}{
    \begin{tabular}{lcccccccgccc}
      \toprule
      Method & MOTA & MOTA & MOTA & MOTA & MOTA & MOTA &  MOTA & {\bf MOTA} & MOTP & Prec & Rec  \\
      & Head & Shou & Elb  & Wri  & Hip  & Knee & Ankl & {\bf Total} & Total& Total& Total\\
      \midrule
      Bbox IoU & 61.7 & 65.5  & 57.3  & 45.7  & 54.3  & 53.1  & 45.7  & 55.2  & 61.5  & 88.1  & 66.5 \\
      Pose PCK & 60.7 & 64.5  & 56.5  & 44.8  & 53.3  & 52.0  & 44.6  & 54.2  & 61.5  & 88.1  & 66.5  \\
      CNN cos-dist & 61.9 & 65.7  & 57.5  & 45.8  & 54.4  & 53.3  & 45.8  & {\bf 55.4}  & 61.5  & 88.1  & 66.5 \\
      All combined & 61.9 & 65.7  & 57.4  & 45.7  & 54.4  & 53.2  & 45.7  & 55.3  & 61.5  & 88.1  & 66.5\\
      \midrule
      LSTM & 54.2 & 58 & 50.4 & 39.4 & 47.4 & 46.6 & 39.8 & 48.4 & 61.4 & 88.1 & 66.5\\
      \bottomrule
    \end{tabular}
    }
  \end{center}
  \caption{{\bf Comparison between different similarity cost criteria}.
  We compare various different hand-crafted and learned cost criterion for the matching stage to generate tracks. Interestingly,
  simple hand-crafted approaches perform very well for the task. We choose to go with the simple bounding box overlap due to low
  computational cost and strong performance.
  }\label{tab:cost}
\end{table}

\begin{table}[t]
  \small
  \setlength\tabcolsep{4pt} 
  \begin{center}
    \begin{tabular}{lr|rrr}
      \toprule
      & Ours & Perfect association & Perfect keypoints & Both \\
      (MOTA) & 55.2 & 57.7 & 78.4 & 82.9 \\
      \bottomrule
    \end{tabular}
  \end{center}
  \caption{
  {\bf Upper bounds:} We compare our performance, with our \textit{potential} performance, if we had the following perfect information. a) Perfect association: We modify the evaluation code to copy over the track IDs from ground truth (GT) to our predictions (PD), after assignment is done for evaluation. This shows what our model would achieve, if we could track perfectly (i.e.\ incurring 0 ID switches). b) Perfect keypoints: We replace our PD keypoints with GT keypoint, where GT and PD are matched using box overlap. This shows what our model would achieve, if we  predict keypoints perfectly.
  c) Finally we combine both, and show the performance with perfect keypoints and tracking, given our detections.
  }\label{tab:upper_bound}
\end{table}

\begin{table*}[t]
  \tableSize{}
  \setlength\tabcolsep{4pt} 
  \begin{center}
    \begin{tabular}{lccccccccgccc}
      \toprule
      Method & Dataset & MOTA & MOTA & MOTA & MOTA & MOTA & MOTA &  MOTA & {\bf MOTA} & MOTP & Prec & Rec  \\
             & & Head & Shou & Elb  & Wri  & Hip  & Knee & Ankl & {\bf Total} & Total& Total& Total\\
      \midrule
      Final model & (Mini) Test v1.0 &  55.9 & 59.0 & 51.9 & 43.9 & 47.2 & 46.3 & 40.1 & {\bf 49.6} & 34.1 & 81.9 & 67.4 \\
      PoseTrack~\cite{iqbal2016pose} & Test (subset)& - & - & - & - & - & - & - & 46.1 & 64.6 & 74.8 & 70.5 \\
    \bottomrule
    \end{tabular}
  \end{center}
  \caption{{\bf Final performance on test set}. We compare our method with the previously reported method on a subset of this dataset~\cite{iqbal2016pose}. 
  Note that~\cite{iqbal2016pose} reports performance at PCKh0.34; the comparable PCKh0.5 performance was provided via personal communication.
  Our performance was obtained by submitting our predictions to the evaluation server. Our model was a ResNet-101 base trained on train+val sets, and 
  tracking was performed at 0.95 initial detection threshold, hungarian matching and bbox overlap cost criterion. This model also out-performed all 
  competing approaches in the ICCV'17 PoseTrack challenge~\cite{posetrack_challenge}.
  }\label{tab:final}
\end{table*}

\begin{figure*}
    \centering
    \includegraphics[width=\linewidth,trim={0 8cm 0 0},clip]{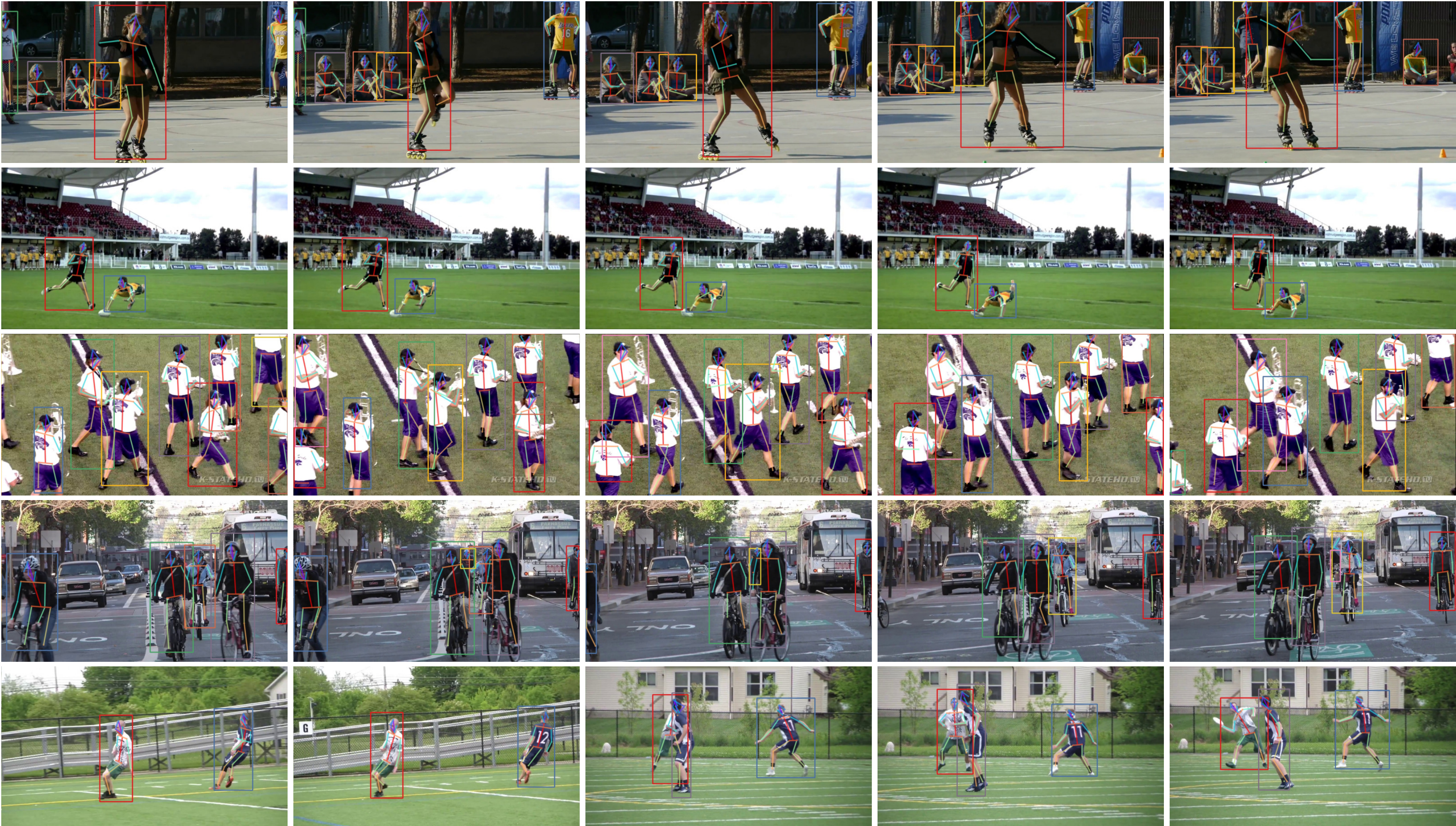}
    \caption{{\bf Sample results.} Visualization of predictions from our two-stage model on the PoseTrack validation set. We show five frames per video, with each
    frame labeled with the detections and keypoints. The detections are color coded by predicted track id. Note that our model is able to successfully
    track people (and hence, their keypoints) in highly cluttered environments. One failure case of our model, as illustrated by the last video clip above,
    is loss of track due to occlusion. As the skate-boarder goes behind the pole, the model loses the track and assigns a new track ID after it recovers
    the person.
    }
    \label{fig:results}
\end{figure*}

{\noindent \bf Upper bounds:}\label{sec:expts:upperbounds}
One concern about the linking stage is that it is relatively simple, and does not handle occlusions or missed detections.
However we find that even without explicit occlusion handling, our model is not significantly affected by it.
To prove this, we compute the {\em upper bound} performance given our detections and given {\em perfect} data association. Perfect data association indicates that tracks are preserved in time even when detections are missed at the frame level. As explained in Table~\ref{tab:upper_bound}, we obtain a small 2.5\% improvement in MOTA ($55.2 \rightarrow 57.5$) compared to our box-overlap based association, indicating that our simple heuristic is already close to optimal in terms of combinatorial matching. As shown in Table~\ref{tab:upper_bound}, the biggest challenge is the quality of the pose estimates. Note that a very substantial boost is observed when perfect pose predictions are assumed, {\em given our detections} ($55.2 \rightarrow 78.4$). This shows that the biggest challenge in PoseTrack is building better pose predictors.
Note that
our approach can be modified to handle jumps or holes in tracks
matching over the previous $K$ frames as opposed to only the last frame, similar to~\cite{pirsiavash2011globally}. This would allow for propagation of tracks even if a detection is missed in $K-1$ frames, at a cost linear in $K$.

{\noindent \bf Comparison with state of the art:}
We now compare our baseline tracker to previously published work on this dataset. Since this data was released only recently,
there are no published baselines on the complete dataset. However, previous work~\cite{iqbal2016pose} from the authors of the challenge reports results on a subset of this data. We compare our performance on the test set (obtained from the evaluation server) to their performance in Table~\ref{tab:final}.
Note that their reported performance in~\cite{iqbal2016pose} was at PCKh0.34, and the PCKh0.5 performance was provided via personal
communication.
We note that while the numbers are not exactly comparable due to differences in the test set used, it helps put our approach in
perspective to the state of the art IP based approaches.
We also submitted our final model to the ICCV'17 challenge~\cite{posetrack_challenge}.
Our final MOTA performance on the full test set was 51.8, and 
out-performed all competing approaches submitted to the challenge. Fig.~\ref{fig:results} shows some sample results of our approach.

\begin{table*}[t]
  \footnotesize
  \setlength\tabcolsep{2pt} 
  \begin{center}
    \resizebox{\linewidth}{!}{
    \begin{tabular}{lllcccccccg|cccccccgccc}
      \toprule
       & Init & Style & mAP  & mAP  & mAP  & mAP & mAP & mAP  & mAP   & {\bf mAP}   & MOTA & MOTA & MOTA & MOTA & MOTA & MOTA & MOTA & {\bf MOTA} & MOTP & Prec & Rec  \\
       & & & Head & Shou & Elbo & Wri & Hip & Knee & Ankl & {\bf Total} & Head & Shou & Elb  & Wri  & Hip  & Knee & Ankl & {\bf Total} & Total& Total& Total\\
      \midrule
      2D & ImNet & & 23.4 & 20  & 12.3  & 7.8 & 16.9  & 11.2  & 8 & 14.8    & 17.5  & 12.9  & -2.4  & -11.6 & 8.2 & -1.6  & -7.6  & 3.2 & 0 & 57.8  & 22.6 \\
      3D & ImNet & center & 27  & 22  & 13.3  & 7.8 & 19.2  & 12.8  & 9.4 & {\bf 16.7}    & 21.5  & 13.3  & -2.1  & -11.7 & 8.4 & -2  & -6.3  & {\bf 4.3} & 10.3  & 56.9  & 24.6 \\
      3D & ImNet & mean & 26.7  & 20.1  & 11.6  & 6.9 & 19.2  & 12.4  & 9.1 & 15.9    & 21.1  & 12.1  & -4  & -14.7 & 8.1 & -2.4  & -7.1  & 3.2 & 10  & 55.6  & 24.6 \\
      \midrule
      2D & COCO & & 28.7  & 25.4  & 17.5  & 10.8  & 24.4  & 17.1  & 11.3  & 19.9    & 24.6  & 20.8  & 11.9  & 4.7 & 17.9  & 11.1  & 2.6 & 14.1  & 5.4 & 73.6  & 24.6 \\
      3D & COCO & center & 32.5 & 30.4  & 19.9  & 12  & 26.6  & 18.7  & 13.5  & {\bf 22.6}    & 27.7  & 24.5  & 12.1  & 4.5 & 18.7  & 11.2  & 3.3 & {\bf 15.4}  & 31.3  & 70.7  & 28 \\
      3D & COCO & mean & 29.3 & 26.4  & 18.2  & 10.4  & 24.9  & 16.8  & 12.1  & 20.4    & 25  & 21.5  & 10.8  & 1.1 & 18  & 9.6 & 2.3 & 13.4  & 15  & 69.7  & 25.7\\
      \bottomrule
    \end{tabular}
  }
  \end{center}
  \caption{{\bf \MODEL{} performance}. 
  We compare our proposed \MODEL{} model with the baseline 2D model that achieves state of the art performance on the PoseTrack challenge. Due to GPU
  memory limitations, we use a ResNet-18 base architecture for both models with frames resized to 256 pixels (this leads to a drop in performance compared to ResNet-50, over
  800px images). Our 3D model outperforms the 2D frame-level model for the tracking task in both MOTA and mAP metrics. We observe slightly
  higher performance with our proposed ``center'' initialization (as opposed to the ``mean'' initialization proposed in~\cite{carreira2017quo}).
  }\label{tab:3dmrcnn}
\end{table*}

{\noindent \bf Run-time comparison:}\label{sec:expts:runtime}
Finally, we compare our method in terms of run-time, and show that our
method is nearly two orders of magnitude faster than previous work.
The IP-based method~\cite{iqbal2016pose},
using provided code 
takes 20 hours for a 256-frame video, in 3 stages: a) multiscale pose heatmaps:  15.4min, b) dense matching: 16 hours \& c) IP optimization: 4 hours.
Our method for the same video takes 5.2 minutes, in 2 stages: a) Box/kpt extract: 5.1 min \& b) Hungarian: 2s, leading to a
237$\times$ speedup.
More importantly, 
the run time for~\cite{iqbal2016pose} grows non-linearly,
making it impractical for longer videos.
Our run time, on the other hand, grows linearly with number of frames, 
making it much more scalable to long videos.

\subsection{Evaluating \MODEL{}}\label{sec:exp:3dmrcnn}

So far we have shown results with our baseline model, running frame-by-frame (stage 1),
and constructing the tracks on top of those predictions (stage 2). Now we experiment with our proposed \MODEL{} model, which naturally encodes temporal context by
taking a short clip as input and produces spatiotemporal tubes of humans with keypoint locations (described
in Sec.~\ref{sec:approach:3dmrcnn}). At test time, we run this model densely in a sliding window fashion
on the video, and perform tracking on the center frame outputs.

One practical limitation with 3D CNN models is the GPU memory usage. Due to limits of the current
hardware, we choose to experiment with a lightweight setup of our proposed baseline model.
We use a ResNet-18 base architecture with an image resolution of 256 pixels.
For simplicity, we experiment with $T=3$ frame clips without temporal
striding, although our model can work with arbitrary length clips.
Our model predicts tubes of $T$ frames, with keypoints corresponding to each box in the tube.
We experiment with inflating the weights from both ImageNet and COCO, using
either ``mean'' or ``center'' initialization as described in Sec.~\ref{sec:approach:3dmrcnn}.
Table~\ref{tab:3dmrcnn} shows a comparison between our 3D Mask R-CNN and the  2D baseline.
We re-train the COCO model on ResNet-18 (without the 256px resizing) to be able to initialize our 3D models.
We obtain a mAP of 62.7\% on COCO minival, which is comparable
to the reported performance of ResNet-50 (64.2\% mAP, Table 6~\cite{he2017mask}).
While the initial performance of the 2D model drops due to small input resolution and shallower model, we see clear gains by using our 3D model on the same
resolution data with same depth of the network. This suggests potentially similar gains over the deeper model as well, once GPU/systems limitations
are resolved to allow us to efficiently train deeper \MODEL{} models.

\section{Conclusion and Future Work}

We have presented a simple, yet efficient approach to human keypoint tracking in videos. Our approach combines the state-of-the-art in frame-level pose estimation with a fast and effective person-level tracking module to connect keypoints over time. Through extensive ablative experiments, we explore
different design choices for our model, and present strong results on the PoseTrack challenge benchmark.
This shows that a simple Hungarian matching algorithm on top of good keypoint predictions is sufficient for getting strong performance for
keypoint tracking, and should serve as a strong baseline for future research on this problem and dataset. For frame-level pose estimation we experiment with both a Mask R-CNN as well as our own proposed 3D extension of this model, which leverages temporal information from small clips to generate more robust predictions. Given the same base architecture and input resolution, we found our \MODEL{} to yield superior results to the 2D baseline. 
However, our 2D baseline requires less GPU memory and as a result can be applied to high image resolutions (up to 800 pixels) with high-capacity models (ResNet-101), which elevate the performance of this simple 2D baseline to state of the art results on the challenging PoseTrack benchmark. 
We believe that as GPU capacity increases and systems become capable splitting and training models across multiple GPUs, there is a strong
potential for \MODEL{} based approaches, especially when applied to high-resolution input and high-capacity base models. We plan to explore those directions as future work.

 \ifcvprfinal{\small
{\noindent \bf Acknowledgements:}
Authors would like to thank Deva Ramanan and Ishan Misra for many helpful discussions.
This research is based upon work supported in part by NSF Grant 1618903, the Intel Science and Technology Center for Visual Cloud Systems (ISTC-VCS), Google, and the Office of the Director of National Intelligence (ODNI), Intelligence Advanced Research Projects Activity (IARPA), via IARPA R\&D Contract No. D17PC00345. The views and conclusions contained herein are those of the authors and should not be interpreted as necessarily representing the official policies or endorsements, either expressed or implied, of ODNI, IARPA, or the U.S. Government. The U.S. Government is authorized to reproduce and distribute reprints for Governmental purposes notwithstanding any copyright annotation thereon.
} \fi

{\small
\bibliographystyle{ieee}
\bibliography{refs}

\begin{thebibliography}{10}\itemsep=-1pt

\bibitem{posetrack_challenge}
Posetrack challenge: {ICCV} 2017.
\newblock \url{https://posetrack.net/iccv-challenge/}.

\bibitem{andriluka20142d}
M.~Andriluka, L.~Pishchulin, P.~Gehler, and B.~Schiele.
\newblock {2D} human pose estimation: {N}ew benchmark and state of the art
  analysis.
\newblock In {\em CVPR}, 2014.

\bibitem{bernardin2008evaluating}
K.~Bernardin and R.~Stiefelhagen.
\newblock Evaluating multiple object tracking performance: the {CLEAR MOT}
  metrics.
\newblock {\em EURASIP Journal on Image and Video Processing}, 2008.

\bibitem{cao2017realtime}
Z.~Cao, T.~Simon, S.-E. Wei, and Y.~Sheikh.
\newblock Realtime multi-person 2d pose estimation using part affinity fields.
\newblock In {\em CVPR}, 2017.

\bibitem{carreira2017quo}
J.~Carreira and A.~Zisserman.
\newblock Quo vadis, action recognition? {A} new model and the kinetics
  dataset.
\newblock In {\em CVPR}, 2017.

\bibitem{chao15hico}
Y.-W. Chao, Z.~Wang, Y.~He, J.~Wang, and J.~Deng.
\newblock Hico: A benchmark for recognizing human-object interactions in
  images.
\newblock In {\em ICCV}, 2015.

\bibitem{LRCN}
J.~Donahue, L.~A. Hendricks, S.~Guadarrama, S.~V. M.~Rohrbach, K.~Saenko, and
  T.~Darrell.
\newblock Long-term recurrent convolutional networks for visual recognition and
  description.
\newblock In {\em CVPR}, 2015.

\bibitem{Feichtenhofer16spatiotemporal}
C.~Feichtenhofer, A.~Pinz, and R.~P. Wildes.
\newblock Spatiotemporal residual networks for video action recognition.
\newblock In {\em NIPS}, 2016.

\bibitem{Feichtenhofer17spatiotemporal}
C.~Feichtenhofer, A.~Pinz, and R.~P. Wildes.
\newblock Spatiotemporal multiplier networks for video action recognition.
\newblock In {\em CVPR}, 2017.

\bibitem{Feichtenhofer16convolutional}
C.~Feichtenhofer, A.~Pinz, and A.~Zisserman.
\newblock Convolutional two-stream network fusion for video action recognition.
\newblock In {\em CVPR}, 2016.

\bibitem{Feichtenhofer17DetectTrack}
C.~Feichtenhofer, A.~Pinz, and A.~Zisserman.
\newblock Detect to track and track to detect.
\newblock In {\em ICCV}, 2017.

\bibitem{Fortman80}
T.~E. Fortman, Y.~Bar-Shalom, and M.~Scheffe.
\newblock Multi-target tracking using joint probabilistic data association.
\newblock 1980.

\bibitem{Girdhar_17b_AttentionalPoolingAction}
R.~Girdhar and D.~Ramanan.
\newblock Attentional pooling for action recognition.
\newblock In {\em NIPS}, 2017.

\bibitem{Girdhar_17a_ActionVLAD}
R.~Girdhar, D.~Ramanan, A.~Gupta, J.~Sivic, and B.~Russell.
\newblock {ActionVLAD}: Learning spatio-temporal aggregation for action
  classification.
\newblock In {\em CVPR}, 2017.

\bibitem{gkioxari15rstar}
G.~Gkioxari, R.~Girshick, and J.~Malik.
\newblock Contextual action recognition with {R*CNN}.
\newblock In {\em ICCV}, 2015.

\bibitem{gurobi}
I.~Gurobi~Optimization.
\newblock Gurobi optimizer reference manual, 2016.

\bibitem{he2017mask}
K.~He, G.~Gkioxari, P.~Doll{\'a}r, and R.~Girshick.
\newblock Mask {R-CNN}.
\newblock In {\em ICCV}, 2017.

\bibitem{He15resnet}
K.~He, X.~Zhang, S.~Ren, and J.~Sun.
\newblock Deep residual learning for image recognition.
\newblock In {\em CVPR}, 2016.

\bibitem{hochreiter1997lstm}
S.~Hochreiter and J.~Schmidhuber.
\newblock Long short-term memory.
\newblock {\em Neural computation}, 1997.

\bibitem{hou2017tcnn}
R.~Hou, C.~Chen, and M.~Shah.
\newblock Tube convolutional neural network (t-cnn) for action detection in
  videos.
\newblock In {\em ICCV}, 2017.

\bibitem{insafutdinov2016articulated}
E.~Insafutdinov, M.~Andriluka, L.~Pishchulin, S.~Tang, B.~Andres, and
  B.~Schiele.
\newblock Articulated multi-person tracking in the wild.
\newblock In {\em CVPR}, 2017.

\bibitem{insafutdinov2016deepercut}
E.~Insafutdinov, L.~Pishchulin, B.~Andres, M.~Andriluka, and B.~Schiele.
\newblock Deepercut: A deeper, stronger, and faster multi-person pose
  estimation model.
\newblock In {\em ECCV}, 2016.

\bibitem{ioffe2015batch}
S.~Ioffe and C.~Szegedy.
\newblock Batch normalization: Accelerating deep network training by reducing
  internal covariate shift.
\newblock In {\em ICML}, 2015.

\bibitem{PoseTrack}
U.~Iqbal, A.~Milan, M.~Andriluka, E.~Ensafutdinov, L.~Pishchulin, J.~Gall, and
  S.~B.
\newblock Pose{T}rack: {A} benchmark for human pose estimation and tracking.
\newblock {\em arXiv:1710.10000 [cs]}, 2017.

\bibitem{posetrack_data}
U.~Iqbal, A.~Milan, M.~Andriluka, E.~Ensafutdinov, L.~Pishchulin, J.~Gall, and
  S.~B.
\newblock Pose{T}rack dataset.
\newblock {PoseTrack / CC INT’L 4.0 / \url{https://posetrack.net/about.php}},
  2017.

\bibitem{iqbal2016pose}
U.~Iqbal, A.~Milan, and J.~Gall.
\newblock Pose-track: Joint multi-person pose estimation and tracking.
\newblock In {\em CVPR}, 2017.

\bibitem{kang2016tcnn}
K.~Kang, H.~Li, J.~Yan, X.~Zeng, B.~Yang, T.~Xiao, C.~Zhang, Z.~Wang, R.~Wang,
  X.~Wang, et~al.
\newblock {T-CNN}: Tubelets with convolutional neural networks for object
  detection from videos.
\newblock {\em arXiv preprint arXiv:1604.02532}, 2016.

\bibitem{kang2016object}
K.~Kang, W.~Ouyang, H.~Li, and X.~Wang.
\newblock {Object Detection from Video Tubelets with Convolutional Neural
  Networks}.
\newblock In {\em CVPR}, 2016.

\bibitem{Karpathy_14}
A.~Karpathy, G.~Toderici, S.~Shetty, T.~Leung, R.~Sukthankar, and L.~Fei-Fei.
\newblock Large-scale video classification with convolutional neural networks.
\newblock In {\em CVPR}, 2014.

\bibitem{kay2017kinetics}
W.~Kay, J.~Carreira, K.~Simonyan, B.~Zhang, C.~Hillier, S.~Vijayanarasimhan,
  F.~Viola, T.~Green, T.~Back, P.~Natsev, et~al.
\newblock The kinetics human action video dataset.
\newblock {\em arXiv preprint arXiv:1705.06950}, 2017.

\bibitem{krizhevsky2012imagenet}
A.~Krizhevsky, I.~Sutskever, and G.~E. Hinton.
\newblock Imagenet classification with deep convolutional neural networks.
\newblock In {\em NIPS}, 2012.

\bibitem{hmdb51}
H.~Kuehne, H.~Jhuang, E.~Garrote, T.~Poggio, and T.~Serre.
\newblock {HMDB}: a large video database for human motion recognition.
\newblock In {\em ICCV}, 2011.

\bibitem{kuhn1955hungarian}
H.~W. Kuhn.
\newblock The hungarian method for the assignment problem.
\newblock {\em Naval Research Logistics (NRL)}, 1955.

\bibitem{lin2017FPN}
T.-Y. Lin, P.~Dollár, R.~Girshick, K.~He, B.~Hariharan, and S.~Belongie.
\newblock Feature pyramid networks for object detection.
\newblock In {\em CVPR}, 2017.

\bibitem{coco_dataset}
T.-Y. Lin, M.~Maire, S.~Belongie, J.~Hays, P.~Perona, D.~Ramanan,
  P.~Doll{\'a}r, and C.~L. Zitnick.
\newblock {COCO Dataset}.
\newblock {COCO / CC INT’L 4.0 / \url{http://cocodataset.org/\#termsofuse}}.

\bibitem{lin2014microsoft}
T.-Y. Lin, M.~Maire, S.~Belongie, J.~Hays, P.~Perona, D.~Ramanan,
  P.~Doll{\'a}r, and C.~L. Zitnick.
\newblock Microsoft {COCO}: Common objects in context.
\newblock In {\em ECCV}, 2014.

\bibitem{mallya16actions}
A.~Mallya and S.~Lazebnik.
\newblock Learning models for actions and person-object interactions with
  transfer to question answering.
\newblock In {\em ECCV}, 2016.

\bibitem{milan2016mot16}
A.~Milan, L.~Leal-Taix{\'e}, I.~Reid, S.~Roth, and K.~Schindler.
\newblock {MOT16}: {A} benchmark for multi-object tracking.
\newblock {\em arXiv preprint arXiv:1603.00831}, 2016.

\bibitem{milan2017online}
A.~Milan, S.~H. Rezatofighi, A.~R. Dick, I.~D. Reid, and K.~Schindler.
\newblock Online multi-target tracking using recurrent neural networks.
\newblock In {\em AAAI}, 2017.

\bibitem{papandreou2017towards}
G.~Papandreou, T.~Zhu, N.~Kanazawa, A.~Toshev, J.~Tompson, C.~Bregler, and
  K.~Murphy.
\newblock Towards accurate multi-person pose estimation in the wild.
\newblock In {\em CVPR}, 2017.

\bibitem{pirsiavash2011globally}
H.~Pirsiavash, D.~Ramanan, and C.~C. Fowlkes.
\newblock Globally-optimal greedy algorithms for tracking a variable number of
  objects.
\newblock In {\em CVPR}, 2011.

\bibitem{pishchulin2016deepcut}
L.~Pishchulin, E.~Insafutdinov, S.~Tang, B.~Andres, M.~Andriluka, P.~V. Gehler,
  and B.~Schiele.
\newblock Deepcut: Joint subset partition and labeling for multi person pose
  estimation.
\newblock In {\em CVPR}, 2016.

\bibitem{Reid79}
D.~B. Reid.
\newblock An algorithm for tracking multiple targets.
\newblock {\em IEEE Transactions on Automatic Control}, 1979.

\bibitem{ren2015faster}
S.~Ren, K.~He, R.~Girshick, and J.~Sun.
\newblock Faster {R-CNN}: {T}owards real-time object detection with region
  proposal networks.
\newblock In {\em NIPS}, 2015.

\bibitem{sadeghian2017tracking}
A.~Sadeghian, A.~Alahi, and S.~Savarese.
\newblock Tracking the untrackable: Learning to track multiple cues with
  long-term dependencies.
\newblock In {\em ICCV}, 2017.

\bibitem{Simonyan_14b}
K.~Simonyan and A.~Zisserman.
\newblock Two-stream convolutional networks for action recognition in videos.
\newblock In {\em {NIPS}}, 2014.

\bibitem{Simonyan14c}
K.~Simonyan and A.~Zisserman.
\newblock Very deep convolutional networks for large-scale image recognition.
\newblock In {\em ICLR}, 2015.

\bibitem{song2017thin}
J.~Song, L.~Wang, L.~Van~Gool, and O.~Hilliges.
\newblock Thin-slicing network: A deep structured model for pose estimation in
  videos.
\newblock In {\em CVPR}, 2017.

\bibitem{ucf101}
K.~Soomro, A.~R. Zamir, and M.~Shah.
\newblock {UCF101:} {A} dataset of 101 human actions classes from videos in the
  wild.
\newblock {\em CRCV-TR-12-01}, 2012.

\bibitem{szegedy2017inception}
C.~Szegedy, S.~Ioffe, V.~Vanhoucke, and A.~A. Alemi.
\newblock Inception-v4, inception-resnet and the impact of residual connections
  on learning.
\newblock In {\em AAAI}, 2017.

\bibitem{tran2015c3d}
D.~Tran, L.~Bourdev, R.~Fergus, L.~Torresani, and M.~Paluri.
\newblock Learning spatiotemporal features with 3d convolutional networks.
\newblock In {\em ICCV}, 2015.

\bibitem{TSN2016ECCV}
L.~Wang, Y.~Xiong, Z.~Wang, Y.~Qiao, D.~Lin, X.~Tang, and L.~{Van Gool}.
\newblock Temporal segment networks: Towards good practices for deep action
  recognition.
\newblock In {\em ECCV}, 2016.

\bibitem{yang2013articulated}
Y.~Yang and D.~Ramanan.
\newblock Articulated human detection with flexible mixtures of parts.
\newblock {\em PAMI}, 2013.

\bibitem{yu2016solution}
S.-I. Yu, D.~Meng, W.~Zuo, and A.~Hauptmann.
\newblock The solution path algorithm for identity-aware multi-object tracking.
\newblock In {\em CVPR}, 2016.

\bibitem{zhou2014learning}
B.~Zhou, A.~Lapedriza, J.~Xiao, A.~Torralba, and A.~Oliva.
\newblock Learning deep features for scene recognition using places database.
\newblock In {\em NIPS}, 2014.

\end{thebibliography}
}
\end{document}